\documentclass[conference]{IEEEtran}
\ifCLASSINFOpdf
   \usepackage[pdftex]{graphicx}
\else
   \usepackage[dvips]{graphicx}
\fi
\usepackage{amsmath}
\usepackage{array}
\usepackage{multirow}

\usepackage{bm}

\begin{document}
\title{Learning Pain from Action Unit Combinations:\\ A Weakly Supervised Approach via Multiple Instance Learning}


\author{\IEEEauthorblockN{Zhanli Chen}
\IEEEauthorblockA{Department of Electrical and\\Computer Engineering\\
University of Illinois at Chicago\\
Email: zchen35@uic.edu}

\and
\IEEEauthorblockN{Rashid Ansari}
\IEEEauthorblockA{Department of Electrical and\\Computer Engineering\\
University of Illinois at Chicago\\
Email: ransari@uic.edu}
\and
\IEEEauthorblockN{Diana J. Wilkie}
\IEEEauthorblockA{College of Nursing\\
University of Florida\\
Email: diwilkie@ufl.edu}}


%Telephone: (800) 555--1212, Fax: (888) 555--1212}


\maketitle

\begin{abstract}

Patient pain can be detected highly reliably from facial expressions using a set of facial muscle-based action units (AUs) defined by the Facial Action Coding System (FACS). A key characteristic of facial expression of pain is the simultaneous occurrence of pain-related AU combinations, whose automated detection would be highly beneficial for efficient and practical pain monitoring. Existing general Automated Facial Expression Recognition (AFER) systems prove inadequate when applied specifically for detecting pain as they either focus on detecting individual pain-related AUs but not on combinations or they seek to bypass AU detection by training a binary pain classifier directly on pain intensity data but are limited by lack of enough labeled data for satisfactory training. In this paper, we propose a new approach that mimics the strategy of human coders of decoupling pain detection into two consecutive tasks: one performed at the individual video-frame level and the other at video-sequence level. Using state-of-the-art AFER tools to detect single AUs at the frame level, we propose two novel data structures to encode AU combinations from single AU scores. Two weakly supervised learning frameworks namely multiple instance learning (MIL) and multiple clustered instance learning (MCIL) are employed corresponding to each data structure to learn pain from video sequences. Experimental results show an $87\%$ pain recognition accuracy with $0.94$ AUC (Area Under Curve) on the UNBC-McMaster Shoulder Pain Expression dataset. Tests on long videos in a lung cancer patient video dataset demonstrates the potential value of the proposed system for pain monitoring in clinical settings.
\end{abstract}

% AFER, Multiple Instance Learning, FACS, Action Unit Combinations, Automated Pain Detection

% \fi
\IEEEpeerreviewmaketitle

\section{Introduction}

Assessing pain is a difficult but important task in clinical settings, which in practice relies on self-report by patients through simple subjective pain assessment measures like visual analog scale (VAS). Research has shown that facial expressions can serve as reliable indicators of pain across human lifespan \cite{craig1992facial} and there is also good consistency of facial expressions corresponding to pain stimuli. The Facial Action Coding System (FACS) is widely used in pain analysis because it provides an objective assessment to score and recognize Action Units (AUs), which represent the muscular activity that produces momentary changes in facial appearance \cite{ekman1997face}. Several studies \cite{deyo2004development}, \cite{prkachin2009assessing} using FACS have identified a collection of core Action Units, which are specific to pain and that occur singly or in combination as summarized in Table I. These results are also confirmed in the study of facial expressions of pain suffered by cancer patients \cite{wilkie1995facial}. Note that although AU27 (mouth stretch) is included in \cite{wilkie1995facial}, we did not include it in this study as it occurs infrequently. Facial expression annotation of videos using FACS is generally performed offline by trained experts who closely examine  the video of a patient's face. A long video is typically divided into multiple subsequences of fixed length duration and AUs are coded at each time step (i.e. each video frame) within the video subsequence. Pain is assessed across the entire sequence based on the occurrence and frequency of pain-related AUs. However the Action Unit coding via human observations is very time consuming, which makes its real-time clinical use prohibitive \cite{lucey2009automatically},\cite{ashraf2009painful}. Therefore, the development of an efficient real-time automated FACS-based pain detection would be a significant innovation for enhanced patient care and clinical practice efficiency.

\begin{table}[ht]
\caption{Action Unit Definition and Pain-Related AU combinations} % title of Table
\centering % used for centering table
\begin{tabular}{|c|c|c|} % centered columns (2 columns)
\hline %inserts double horizontal lines
AU & Description & Pain-Related Combinations\\ [0.5ex] % inserts table heading
\hline\hline % inserts single horizontal line
4 & eye brow lower & 6/7 \\ % inserting body of the table
6 & cheek raiser & 20\\
7 & eye lid tightener & 4+6/7/43\\ % inserting body of the table
9 & nose wrinkler & 4+9/10\\
10 & upper lip raiser & 4+26\\ % inserting body of the table
20 & lip stretcher & 9/10+26\\
26 & jaw drop & \\
43 & eyes closed &  \\[1.5ex] % [1.5ex] adds vertical space
\hline %inserts single line
\end{tabular}
\label{table:nonlin} % is used to refer this table in the text
\end{table}

Progress in computer vision and machine learning (CVML) techniques over time has led to significant development of general Automated Facial Expression Recognition (AFER) systems, although limited effort has been reported for its application in pain analysis. One major challenge is the difficulty in establishing a comprehensive annotated dataset with sufficient examples of pain-related expressions. Most existing video datasets containing pain-related facial expressions are developed for targeted studies. These datasets are typically not publicly accessible. They are mostly small in size and lack sufficient diversity to train a robust automated system in general. One development that has facilitated the research on spontaneous expressions recognition is the introduction of UNBC-McMaster Shoulder Pain Archive, which is the only publicly available comprehensive pain-oriented facial expression dataset. This dataset contains recordings of subjects experiencing shoulder pain in a clinical setting with complete labeling at both the frame level (AUs) and the sequence level (OPI, VAS, etc.), and it has been widely employed by FACS-based AFER research as a standard dataset for performance evaluation. On the other hand, shoulder pain involved in UNBC-McMaster dataset is acute pain, and we are not aware of any research suggesting that this dataset is beneficial to study other type of pain, such as chronic pain caused by cancer.

An important observation that motivated our research is that existing research on FACS-based automated AU recognition focuses on detection of single AUs. We note that pain-related AUs occur in conjunction with other AUs to form combinations irrelevant to pain. Therefore inference based only on occurrence of individual AUs is not sufficient for pain identification. While the ground truth of facial expressions and action units in video databases is available at frame level, the ground truth about pain is typically available at sequence level and only via self-report, which is an example of 'weakly labeled' data. In early attempts of automated pain analysis, proposed approaches \cite{lucey2009automatically}\cite{ashraf2009painful} employed an averaging paradigm by assigning the sequence label of pain occurrence to each frame and training a support vector machine (SVM) on the frame-level label. Pain is declared to occur in a video if the average output of frame-level pain score exceeds a threshold. However, pain-related frames may constitute a small fraction of all frames in a long video. In this case averaging the output score could therefore severely attenuate the signal of interest. Recent research \cite{sikka2014classification} suggests that video-based pain detection can be formulated as a weakly supervised learning problem, and multiple instance learning (MIL) is an effective machine learning tool to handle this problem. In the study \cite{sikka2014classification}, a binary pain classifier is trained directly via the high-dimensional features extracted from video frames without going through the AU coding procedure. Although encouraging results are reported from experiments on UNBC-McMaster dataset, this setting is vulnerable to performance degradation for trans-dataset application, due to interference from person-specific features and demographic variations encoded in the the high-dimensional features.

% pain manifest through facial expressions

In a commonly used procedure for the manual detection of pain, FACS-certified coders first perform AU coding for every video frame and then infer the sequence level pain label from the occurrence and frequency of pain-related AU combinations. In general, AU occurrence is highly correlated with the appearance of pain in facial expressions, and the reliability of pain detection strongly depends on the accuracy of AU coding. However, facial expressions of pain are more likely to appear when the pain intensity level is high or when intensity of pain changes to a higher level. Due to the sparsity of pain expressions, video captured in clinical settings usually lacks sufficient positive samples to train a reliable pain classifier by learning a direct mapping between high-dimensional facial features and self-report pain labels. On the other hand, state-of-the-art AFER systems can be effectively trained on millions of online image \cite{mcduff2013affectiva}, which possess sufficient adaptability to all kinds of video datasets including pain-oriented ones. Inspired by these observations, we propose a new FACS-based automated pain detection framework comprised of two independent machine learning networks to infer the presence of Action Unit (AU) combinations that signify pain. The two networks are trained independently and are linked by two proposed novel AU-based data structures – compact or clustered- which are created by mapping single AU measurements per frame into pain-related low-dimensional feature vectors representing AU combinations. The first machine learning network is a generic AU detector that takes advantage of current progress in computer vision and machine learning techniques to handle challenges from person-specific variation and natural illumination characteristic of clinical settings. The second machine learning network, called the Multiple Instance Learning (MIL), solves pain detection as a weakly supervised learning problem and performs analysis in the low-dimensional AU combination feature space. This decoupled structure of two independent machine learning networks facilitates data fusion from different pain-oriented video datasets and helps to develop a commercial, robust, and generic automated pain analysis system. The decoupled structure mimics the strategy used by human coders in clinical settings. To our best knowledge, this is the first work on automated pain analysis based on a complete set of pain-related AU combinations.

The rest of the paper is organized as follows, Section 2 provides a review of the literature on automated pain detection and analysis. Section 3 gives a brief overview of two pain-oriented datasets involved in this work: the UNBC-McMaster dataset (acute shoulder pain) and Wilkie's dataset (chronic cancer pain). Section 4 presents the pain detection framework in a decoupled structure, with focus on the feature representation based on AU combinations and corresponding learning tools (MIL and MCIL). Section 5 provides a demonstration of the advantages of the proposed framework based on the results of testing our method on both datasets described in Section 3. Finally Section 6 presents the conclusion of the work.

%Recent research[] propose to   machine learning weakly supervised problem
%
%Mapping between AU and pain, facial expressions and AU, but not pain and FE, which is suitable to be described by BOW
%
%Contributions long video
%
%AFER for general AU recognition. Spontaneous AU low occurrence for social, non pain applica
%The effort for pain recognition based on spontaneous AU detection is limited
%pain recognition facial expressions can be modeled as a weakly supervised machine learning problem
%ground truth is based on VAS, not available on frame level in clinical settings.
%existing work stop at single AU
%Acute pain is studied more than chronic pain cancer pain, dataset is limited
%Most research is based on short videos, clinical applications requires pain monitoring
%Our contribution,
%Decouple the problem, AFER based on state of the art solution, Emotient, can be replaced by other methods
%learning pain from simple features generated from AU combination
%focus on extraction of relations between AUs and  pain, test on Wilkie's dataset that can not be trained a stand alone system
%test on long videos
%imitate human coder decision scoresheet
%high efficiency

\section{Related Past Work On Pain Detection}
In the past decades, significant progress in computer vision and machine learning techniques (CVML) has boosted the development of the AFER system. With increasing demand for facial expression applications, the focus of AFER research shifted from posed expressions obtained in a controlled setting to spontaneous expressions evoked in a natural settings. Further details can be found in survey papers \cite{bousmalis2013towards}\cite{corneanu2016survey}\cite{zeng2009survey}. A robust AFER system is capable of handling perturbations due to demographic difference, environment variation, and rigid motions, and is applicable to different datasets without a need for retraining. Emotient \cite{littlewort2011computer} and Affectiva \cite{mcduff2013affectiva} are two well-known examples of state-of-the-art AFER systems which are commercially available as platforms that facilitate the development of applications based on facial expression. However, the output of such systems is more appropriate as an intermediate result in the form of general AU scores and further processing and customization is required for advanced applications including pain analysis. On the other hand, while popular CVML tools are widely used in most AFER research, very limited effort are reported on exploring machine learning based pain interpretation from facial expressions.  Ashraf et al. \cite{ashraf2009painful} studied the UNBC-McMaster dataset and proposed three feature types that are extracted from the Active Appearance Model (AAM) to train SVM pain classifiers. They used an averaging scheme to generate sequence-level labels. In their subsequent research\cite{lucey2011painful}\cite{lucey2012painful}, the same set of features are used to train a binary classifier for each single pain-related AU at frame level together with a pain intensity classifier at sequence level using the OPI labels. Chen \emph{et al} \cite{chen2012automated} used a simple rule-based method to model temporal dynamics of AUs to study pain of patients suffering from lung cancer in the Wilkie's dataset. Sikka \emph{et al } \cite{sikka2015automated} employed a CVML-based model to assess pediatric postoperative pain on a video dataset of neurotypical youth. A total of $14$ single AUs are extracted under $3$ statistics to form a $42$-dimensional descriptor for each pain event which serves as the input to logistic regression models of both binary pain classification and pain intensity estimation. Later, Sikka \emph{et al } \cite{sikka2014classification} modeled video sequences from UNBC-McMaster dataset with the Bag of Words representation and applied Multiple Segment Multiple Instance Learning (MS-MIL) for jointly detecting and localizing frames containing pain using sequence-level ground truth (OPI). We note three shortcomings observed in previous research : (1) AU recognition and pain recognition are treated as separate problems which are handled by an AU classier and a pain classifier respectively, without giving sufficient attention to their relationship; (2) pain analysis is more focused on single AU detection rather than pain relevant AU combination; (3) The difficulty in having access to video datasets with chronic pain. And also insufficient effort on developing advanced comprehensive automated analysis tools targeted to clinical applications.

%PSPI, sakkie[] pediatric pain, MIL pantic pain intensity  To summarize, modeling temporal information, independent pain classifier,
%including FACS based AU labeling. universal
%Advanced techniques and adequate training has made system robust to illuminations, universal applicable, robust. State of the art Emotient and Affectiva, commercialized software. Good platform for extensive research. Details refer to survey
%Research on pain recognition, first Ashrif et al. Single AU recognition including our early work. Pediatric pain research. MIL learning in Pain. Pantic AU and AU combination work.
%
%applications: human computer interaction, video
%surveillance, forensic applications, criminal investigations,

\section{Available Pain Expression Datasets}

The UNBC-McMaster Shoulder Pain Expression Archive Dataset contains $200$ video sequences captured from patients suffering from shoulder pain and spontaneous facial expressions are triggered by moving their affected and unaffected limbs. All frames are coded by certified FACS coders for $10$ single pain-related AUs and the frame-level pain score is rated by the Prkachin and Solomon Pain Intensity (PSPI). A sequence-level pain label is assigned by self-reported Visual Analog Scale (VAS) and Observer-rated Pain Intensity (OPI). In addition, $66$-point facial landmarks from Active Appearance Model (AAM) are also provided for each frame to facilitate the development of a user-customized AFER system. It is the only publicly available pain-oriented facial expression video dataset in which the spontaneous facial expressions are evoked solely by acute pain.

One major distinguishing feature of this study is that in addition to the UNBC McMaster dataset we also conduct research on a unique dataset created by D. Wilkie \cite{wilkie1995facial}, containing videos of $43$ patients suffering from chronic pain caused by lung cancer. The videos are captured in natural settings in the subjects' homes. The patients were required to repeat a standard set of randomly ordered actions as instructed such as sit, stand up, walk, and recline, in a $10$-minute video with a camera focused on the face area to record their facial expressions. Each video was partitioned into $30$ equal-duration of $20$ seconds per subsequence. The subsequences were reviewed and scored for 9 AUs possibly occurring in combination by three trained human FACS coders independently, and the results were entered in a scoresheet that served as ground truth. Pain was scored in one video subsequence if at least two coders agree with each other on the occurrence of a set of specific AU combinations listed on the scoresheet. The intensity of pain for the entire video was measured on the total number of subsequences that were associated with pain labels. Due to the issue in illumination and video quality degradation, only $1164$ out of the total $1290$ video subsequences were amenable for processsing by Emotient and of these about $600$ subsequences are suitable for pain analysis.

In this paper, we use the UNBC-McMaster dataset to train a MIL based pain detector and then test it on Wilkie's video dataset \cite{wilkie1995facial} with chronic cancer pain. We hypothesize that the pain detector can capture the prevalent feature of pain from the low-dimensional feature vectors of AU combinations that we introduce in this paper.

%Wilkie Lung Cancer Patient Demographic

\section{Proposed Automated Pain Detection Framework}
Pain detection is one of the applications of spontaneous facial expression recognition. It is very beneficial for patient care to develop a pain detection system for clinical settings if we take advantage of existing progress on spontaneous AFER research. Recent state-of-the-art generic AFER systems \cite{imotions2016}\cite{mcduff2013affectiva} have proved to be robust to context variations, especially illumination and rigid motions, and have shown good performance on new datasets with different demographics. This is a key motivation for us to adopt the commercialized Emotient system in the front end of our proposed method. On the other hand, the generic system is not optimized for application to pain recognition，so that deterministic decisions made directly from the AU scores with a preset threshold could cause a high false alarm rate. Therefore we investigate a framework that uses a second layer of machine learning to refine the single AU scores and give a more reliable prediction of pain.

The proposed automated pain detection system is comprised of two-level machine learning systems, an Automated Facial Expression Recognition (AFER) system that computes frame-level confidence scores for single AUs (shown in the left side of Fig.1.) and a MIL system that performs sequence-level pain prediction based on contributions from a pain-relevant set of AU combinations (shown in the right side of Fig.1.). MIL is well-suited to handle the 'weakly labelled' pain data that can be conveniently represented by a bag of word (BOW) structure, and details will be covered in the following chapters. AU combinations are described by two distinct low-dimensional novel feature structures : the compact structure encodes all AU combinations of interest into a single vector, which is analyzed in the MIL framework; the clustered structure uses a sparse representation to encode each AU combination separately followed by analysis in the multiple clustered instance learning (MCIL) framework, which is an extension of MIL. To our knowledge, this is the first work that applies MCIL in facial expression-related research. Both machine learning systems are trained on the UNBC-McMaster dataset independently with frame-level and sequence-level labels, as illustrated by the three circular blocks in the center of Fig.1.

%\begin{figure*}
%  \centering
%  % Requires \usepackage{graphicx}
%  \includegraphics[width=4.5in]{fig_1_framework.png}\\
%  \caption{Pain Detection Framework}\label{1}
%\end{figure*}

\begin{figure}
  \centering
  \includegraphics[width=3.5in]{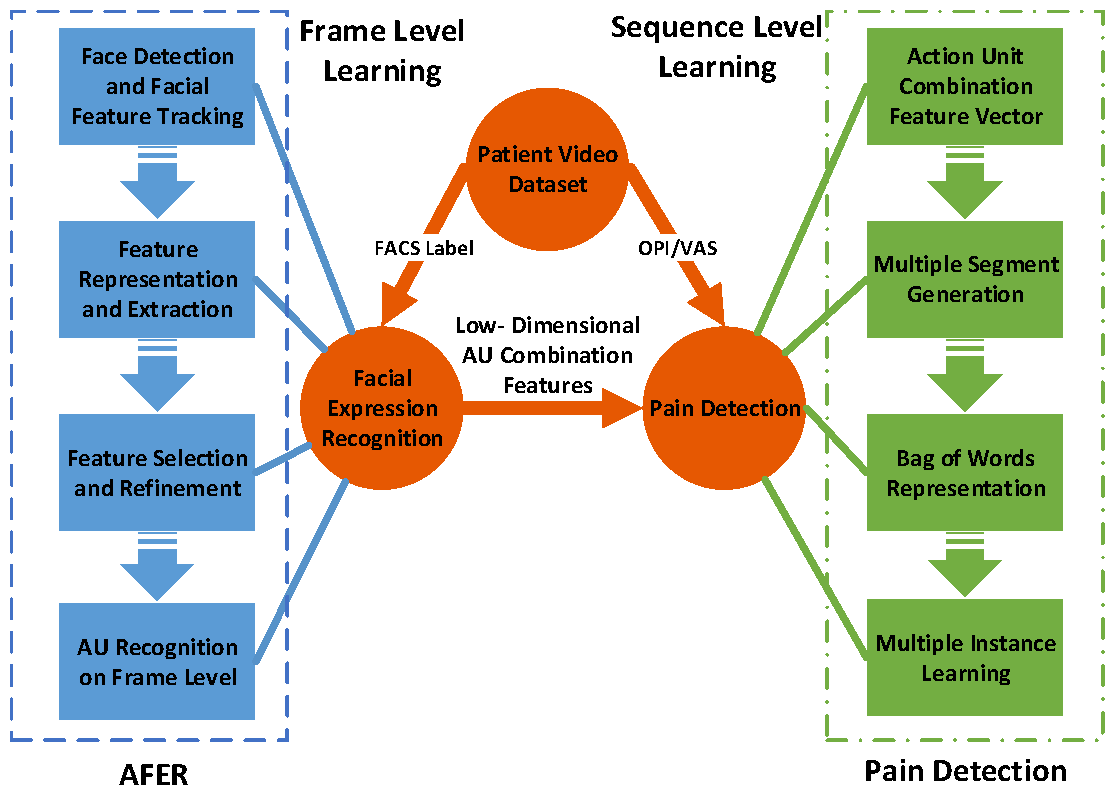}\\
  \caption{The Decoupled Pain Detection Framework}\label{1}
\end{figure}

AU coding relies on observable facial muscular movements or facial expressions, whereas pain is more like a latent variable and is not always manifested itself in facial expressions, especially in the scenario of chronic pain. Due to the limited number of positive labeled samples, it is often difficult to learn a reliable mapping between pain and high-dimensional facial features in a direct manner. On the other hand, the decoupled framework utilizes AU codings as the intermediate results and performs pain analysis in a low-dimensional AU score space, which would be more justifiable and efficient than the direct method. In addition, the decoupled framework alleviates the difficulty in training large scale dataset-specific pain datasets by facilitating data fusion from different pain-labelled video datasets, which may potentially lead to a generic pain detector for analyzing multiple types of pain.

%AU feature vector
%Pain

\subsection{Automated Facial Expression Recognition }
An AFER system can be conveniently described with four key blocks consisting of face detection, feature representation, feature selection, and classification, as shown in Fig.1. The first block identifies face area with a rectangular boundary box in every video frames. The second block aligns the detected face areas and employs various descriptors to extract features from the facial images. The third block selects features most relevant to the non-rigid motions caused by facial expressions, and applies dimension reduction techniques to compress the feature vector to a size that is tractable for the classifier. The fourth block contains a set of one-versus-all classifiers that are trained on the refined feature vectors for each AU of interest, and the output could be either binary decisions or soft scores that reflect the probability or confidence about the target AUs. Existing research \cite{bartlett2006automatic}\cite{koelstra2010dynamic}\cite{lucey2011painful}\cite{jiang2011action} on spontaneous facial expression recognition shows AFER systems are highly customizable, and more blocks could be added to this framework to boost performance depending on the application.

One example of such an AFER system is the computer expression recogntion toolbox (CERT) \cite{littlewort2011computer}, which is the core of the Emotient \cite{imotions2016} system. In the system setup, face detection is based on an extentsion of classic Viola-Jones approach. Ten facial feature points are tracked using GentleBoost and the detected face area is aligned to a canonical template patch through an affine warp estimated from the feature positions. A Gabor filter bank is then applied to extract features in $8$ directions and $9$ spatial frequencies and the filter outputs are concatenated into a single feature vector. The feature vector is then fed into separate linear support vector machines (SVM) for individual AU recognition.

We use Emotient to track and label a set of AUs $\{4,6,7,9,10,20,26,43\}$, that are commonly used in most pain-oriented research. The processing results are represented by the flow of Evidence numbers, where an Evidence number, ranging between $-2$ to $2$ \cite{imotions2016}, is assigned to each AU on every frame. The Evidence output for an expression channel represents the odds in logarithm (base $10$) scale of a target AU being present. For example: Evidence = $2$ ($10^2=100$) means the observed expression is $100$ times more likely to be categorized by an expert human coder as target AU and Evidence = $0$ means the chances that the expression to be categorized by an expert human coder as target AU or not are equal. The Evidence scores can be conveniently transformed to Probability measurements by the following equation:
\begin{equation}
  Probability = \frac{1}{1+10^{-Evidence}}
\end{equation}
The probability measurements derived from the Evidence scores are used as the features of pain analysis framework.  The evidence/probability score profile of every AU can be viewed as a $1D$ time domain signal.

% why Evidence measure metric is more suitable? refer to emotient help page

\subsection{Action Unit Combination Encoding }
% Elaborate more about Wilkie's research strategy on AU coding
% Our system employs the set of AUs in Table 1. to form $11$ pain-related AU combinations as defined in Wilkie's dataset. These AU combinations are grouped into 6 clusters, and a cluster is checked if any member AU combinations in it is scored by human coders.
\subsubsection{\textbf{Compact Structure Vs. Clustered Structure}}
In practice, Action Units $6/7$ are the most frequently observed pain-related AUs in FACS coding \cite{prkachin2009assessing}. Multiple AU combinations could also be activated in a video segment for pain evalutation. Furthermore, the Evidence number produced by AFER suggests uncertainty about AU coding, where higher uncertainty could frequently be associated with spontaneous facial expressions due to their low intensity. Hence when we define feature vectors based on AU combinations, it is important to have a comprehensive characterization in terms of frequency extent of individual AU contribution, and confidence of measurement for each AU combination. Therefore, in the task of designing feature vectors for pain analysis, we not only consider the activation of individual AU combination but also take into account the correlation among activation of multiple AU combinations. Two different feature vector structures based on AU combination scores, which are referred to as compact or clustered, are proposed as follows,

\textbf{Compact Structure}: Let $A=\{4,6,7,9,10,20,26,43\}$ be the set of single pain-related AUs. The AU combination feature vector for frame $i$ of a video sequence $S$ is an $11$ dimensional column vector
\begin{equation*}
\begin{aligned}
  v_{i}=&[P_{AU_6}\, P_{AU_7}\, P_{AU_{20}}\, P_{AU_{4\oplus6}}\, P_{AU_{4\oplus7}}\, P_{AU_{4\oplus43}} \\  & P_{AU_{4\oplus9}}\, P_{AU_{4\oplus10}}\, P_{AU_{4\oplus26}}\, P_{AU_{9\oplus26}}\, P_{AU_{10\oplus26}}]^T
\end{aligned}
\end{equation*}
 in which each entry of the feature vector is the probability estimate of corresponding AU or AU combination and the $\oplus$ operator denotes the co-occurrence of AUs. The probability estimate of the combination $AU(i\oplus j)$ depends on the smaller probability between $AU_i$ and $AU_j$ so that $P_{AU_{i\oplus j}}=min(P_{AU_{i}},P_{AU_{j}}),\forall i,j\in A$. As a result, the pain information about frame $i$ is conveyed by the probability measurement of all pain-related AU combinations that are compressed in a single low-dimensional vector, as shown in Fig. 3(a)
\begin{figure}
  \centering
  \includegraphics[width=2.5in]{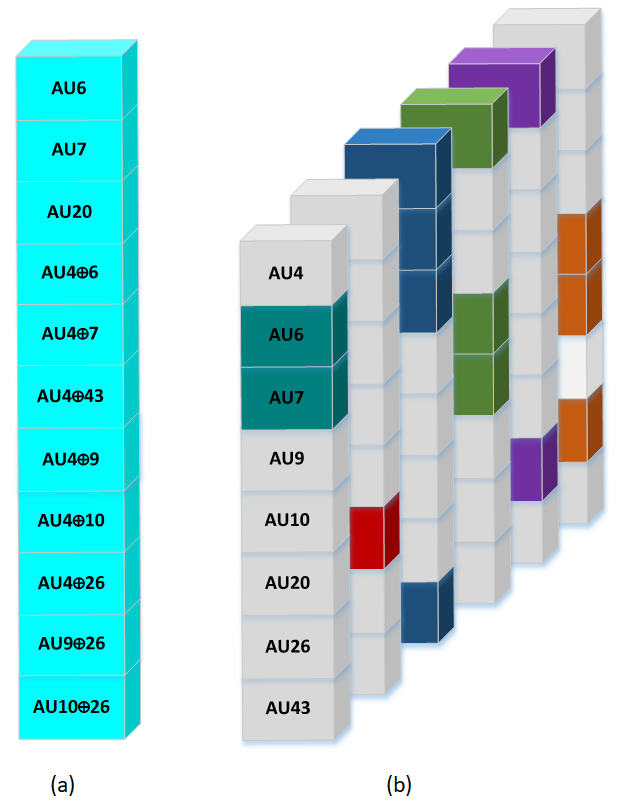}\\
  \caption{AU combination structure: (a)Compact Structure, (b)Clustered Structure}\label{*}
\end{figure}

\textbf{Clustered Structure}: Here we follow Wilkie's coding strategies \cite{wilkie1995facial} to group the $11$ pain-related single AU and AU combinations into clusters according to two criteria, 1) there is a common AU shared by the combinations in the cluster and 2) the AU combinations within a cluster are actuated in adjacent area on the face. Six clusters are formed in this way, including $\{AU_{6/7}\},\{AU_{20}\},\{AU_{4\oplus6/7/43}\},\{AU_{4\oplus9/10}\},\{AU_{4\oplus26}\}$ and $\{AU_{9/10\oplus26}\}$. The AU combination feature representation of a video frame $i$ under the clustered structure is composed of an $8\times6$ matrix, where the column $j$ is highlighted by all the single AUs involved in the combinations belonging to cluster $j$, where $j\in\{1,2,3,4,5,6\}$. The non-zero entry for column $j$ is the probability measurement of the highlighted single AU in cluster $j$ and all the remaining entries of the feature matrix are set to zero, which results in a sparse representation of features, as shown in Fig. 2(b).

\subsubsection{\textbf{Bag of Words representation}} A patient's self-report is still the golden rule for pain evaluation in patient care. A pain label is commonly available for video segment but not for every single frame. Such a situation is encountered frequently in computer vision since it is easier to obtain group labels for the data rather than individual labels, and is known as 'weakly supervised' problem. On the other hand, although temporal dynamics of spontaneous facial expressions have good intra-dataset consistency, it could vary significantly under clinical settings depending on the level and type of pain a patient is suffering. While AUs evoked by acute pain last less than a second, those evoked by chronic cancer pain could last for minutes. Hence conventional temporal modeling with fixed moving window or a preset duration parameter\cite{lucey2009automatically}\cite{koelstra2010dynamic} is inadequate to handle practical applications. To address the challenges from weakly labeled data and complicated temporal dynamics, we employ the bag of word (BOW) representation as suggested in \cite{sikka2012exploring}\cite{sikka2014classification}.

A video sequence $S_{i}$ in a dataset can be represented as a bag that contains a number of segments generated from $S_{i}$. The bag is defined as $\{s_{ij}\}{_{j=1}^{N_{i}}}$， where $s_{ij}$ is the $j$th segment in the bag containing $N_{ij}$ frames i.e. $s_{ij}=\{f{_i^{m_j}},f{_i^{m_j+1}}, \ldots, f{_i^{N_{ij}-m_j+1}} \}$. $s_{ij}$ contains only contiguous frames and $N_{ij}$ is the total number of frames in $s_{ij}$ taken from $S_{i}$ and $f{_i^{m_j}}$ is the $m_j$th frame in $S_{i}$ and the $1$st frame in ${s_{ij}}$. The bags are then associated with the label of sequence $S_{i}$ as $B=\{S_i,y_i\}{_{i=1}^N},\;y_i\in\{-1,\,1\}$， which defines two kinds of bags, positive and negative. A positive bag contains at least one positive instance, while a negative bag contains no positive instances. Adopting this representation for the pain detection problem, a positive bag refers to a video sequence that contains pain-related facial expressions, and a negative bag refers to a video sequence not containing any pain-related facial expression. Practically, pain-related AU temporal segments occupy only a small portion of the entire video sequence. The sparsity of positive training samples fits well in the context of BOW structure, which is another motivation to adopt this type of data structure. It takes three steps, \textbf{S1}-\textbf{S3} as described below, to generate a BOW representation from the feature space.

\textbf{(S1) Feature Extraction at Frame Level}: Define a mapping $\phi_{F}:R^{m\times n}\rightarrow R^d$ as the feature extraction process on frame level that maps a frame of size $m\times n$ in image space to a $d$-dimensional feature vector, where $d==11$ for the compact situation. In the case of clustered structure, the mapping is defined as $\phi_{F}:R^{m\times n}\rightarrow R^D$, where $R^D$ refers to the space of $8\times6$ sparse feature matrices. Feature vectors are typically of very high dimension in existing unified framework. However, $R^d$ or $R^D$ in our case is simply the low-dimensional AU combination feature vector space in the proposed decoupled pain detection framework.

\textbf{(S2) Multiple Segment Generation}: The instances in a bag are video segments containing consecutive frames belonging to the sequence. The bound of each segment can be generated conveniently in two ways. A typical way is to run overlapping temporal scanning windows at multiple scales known as Sc-Wind. A parallel way is clustering the frames in a sequence using normalized cuts (Ncuts). Each element of the weight matrix of Ncut algorithm is obtained by a similarity measure between frames $f{_i^u}$ and $f{_i^v}$ in sequence $i$ measured by
\begin{equation}
\begin{split}
  W(u,v)= & \exp{(-|\frac{\phi_{F}(f{_i^u})-\phi_{F}(f{_i^v})}{\sigma_f}|^2)} \\
    & +\exp{(-|\frac{t_u-t_v}{\sigma_t}|^2)}
\end{split}
\end{equation}
where $t_u$ refers to frame index of $f{_i^u}$, and $\sigma_f$ and $\sigma_t$ are constants selected for feature domain and time domain respectively. Details of Ncuts are provided in\cite{shi2000normalized} \cite{sikka2014classification}.

\textbf{(S3) Feature Representation at Segement Level}: The feature representation of a video segment is denoted by the mapping $\phi_{S}:S\rightarrow R^d$ that transform video segment in sequence space $S$ to a $d$ dimensional feature vector. This mapping is specified by a max-pooling strategy from the feature representation of all the frames in the segment as:
\begin{equation}
  \phi_S(s_{ij})=\mathop{max}_k(\phi_{F}(f_i^k)\,|\,f_i^k\in s_{ij})
\end{equation}
The instance in a bag is now represented by a single feature vector with the same dimension $d$ as the frame-level feature vector. After associating the pain label to the bag, a multiple instance learning(MIL) framework can be trained on the BOW data for automated pain detection.

 % Multiple instances video segmentation in the bag temporal precision

\subsection{Multiple Instance Learning }
The general idea for solving machine learning problems is to establish a classifier and optimize it with respect to a loss function. Viola \emph{et al.}\cite{viola2005multiple} first solved the MIL problem with a boosting framework, which is known as MILboost, and discussed its application in object detection from images. In this section, we give a brief overview of MILboost and how it can be customized for pain detection. The decision on the presence of pain is based on the probability of bags been positive. The posterior probabilities of bags and instances are defined as:
\begin{align}
  p_i & =\mathcal{P}(y_i=1|S_i) \\
  p_{ij} & =\mathcal{P}(y_i=1|s_{ij})
\end{align}
The only available ground truth is the label of the bag, and all the instances in a bag carry the same label as the bag.

A classifier $H_T:R^d \rightarrow R$  is trained on the feature vectors of instances, and a posterior probability is assigned to each instance based on the classifier output, $s.t.$
\begin{equation}
  p_{ij} = \sigma(H_T(\phi_s(s_{ij})))
\end{equation}
where $\sigma()$ is a sigmoid function $s.t.$
\begin{equation}
  \sigma(x) = \frac{1}{1+\exp(-x)},\:\forall x\in R
\end{equation}

The loss function is defined by negtive log-likelihood, which is the same as that used in the logistic regression problem:
\begin{equation}
   \mathcal{L}=-\sum_i^N (r_i*p_i+(1-r_i)(1-p_i))
\end{equation}
where $r_i=1$ if $y_i=1$ and $r_i=0$ if $y_i=-1$.
Since a positive bag contains at least one positive instance, the probability of bag to be positive $p_i$ depends on the probability of the instance that is most likely to be classified as positive, i.e.
\begin{equation}
  p_i=\mathop{max}_j(p_{ij})
\end{equation}

The MILboost uses the boosting procedure to construct a strong classifier $H_T(s_{ij})$ by iteratively combining a set of weak classifiers $h_t(s_{ij})$ as,
\begin{equation}
  H_T(s_{ij})=\sum_{t=1}^T \alpha_{t} h_t(s_{ij})
\end{equation}
where $H_T$ denotes the classifier constructed in the $T^{th}$ iteration and all weak classifiers are the same type of learners that are generated from space $\mathcal{H}$. Note $H_T(s_{ij})$ and $h_t(s_{ij})$ are simplified notations of $H_T(\phi_s(s_{ij}))$ and $h_t(\phi_s(s_{ij}))$ respectively, and similar notation will be used in the following derivations.

The boosting algorithm updates the weight of instances at the end of each iteration by taking the gradient of the loss function $\mathcal{L}$ $w.r.t$ $H_T(s_{ij})$
\begin{equation}
  \omega_{ij}=-\frac{\partial \mathcal{L}}{\partial H_T(s_{ij})}
  =-\frac{\partial\mathcal{L}}{\partial p_{i}}\frac{\partial p_{i}}{\partial p_{ij}}\frac{\partial p_{ij}}{\partial H_T(s_{ij})}
\end{equation}
The instance weights are then normalized as $\omega_{ij}'=\frac{|\omega_{ij}|}{\sum_{ij}|\omega_{ij}|}$.  Misclassified instances will be assigned higher weights and a weak classifier $h_{T+1}(s_{ij})$ is trained on the reweighted data and added to $H_{T+1}$ in the $(T+1)^{th}$ iteration
\begin{equation}
  h_{T+1}=\arg\max_{h\in\mathcal{H}}\sum_{i}^{N}\sum_{j}^{N_i}\omega_{ij}'h(s_{ij})
\end{equation}
where $N$ is the total number of bags and $N_i$ is the number of instances in the $i$th bag. However, since the $\max$ function is not differentiable, a soft-max function $g$ is used as an approximation. Define a set $ \textbf{p}_i =\{p_{i1}\, p_{i2}\, ...\, p_{iN_i}\}$, then the softmax function with respect to the subscript $j,\,j\in\{1,2,...,N_i\}$ is given by:
\begin{equation}
  p_i=g(\textbf{p}_i)=g_j(p_{ij})\approx\mathop{max}_j(p_{ij})
\end{equation}
Among all the options for soft-max function, the generalized mean (GM) is a preferable model as suggested by past research \cite{sikka2014classification}. For the instances$\{s_{ij}\}{_{j=1}^{N_{i}}}$ in the bag of $S_i$, the GM approximation is given by:
\begin{equation}
  g_{GM}(\textbf{p}_i)=g_j(p_{ij})=(\frac{1}{N_i}\sum_j p_{ij}^u)^{\frac{1}{u}}
\end{equation}
where $u$ is the parameter that controls sharpness and accuracy in GM model $s.t.$ $g_{GM}(\textbf{p}_i)\rightarrow \max_j(p_{ij})$ as $u\rightarrow \infty$. Now the gradient of the GM soft-max is given by:
\begin{equation}
  \frac{\partial p_{i}}{\partial p_{ij}}=p_i\frac{p_{ij}^{(u-1)}}{\sum^{N_i}_{s=1} p_{is}^u}
\end{equation}

\subsection{Multiple Clustered Instance Learning }
In the compact structure feature settings, scores of all AU combinations are encoded in one feature vector, which can be conveniently handled by the original MIL framework. However, it may be desirable to distinguish the contribution of individual AU combinations for more precise analysis on different types of pain. Practically, the clustered representation is a more natural way that is used by human coders and a positive decision on any of the clusters is sufficient to identify pain. Multiple Clustered Instance Learning (MCIL) proposed by Xu \emph{et al.}\cite{xu2014weakly} is an extension of MIL that was proposed to provide patch-level clustering of $4$ subclasses of cancer tissues, and facilitates both image-level classification and pixel level segmentation (cancer vs. non-cancer). Based on the structural similarity of the problems, we have adapted the MCIL framework to handle the clustered feature structure for pain recognition.

MCIL assumes there are $K$ clusters in a positive bag and an associated hidden variable $y_{ij}^k\in\{-1,1\}$ that indicates whether the instance $s_{ij}$ belongs to the $k$th cluster. An instance could be considered positive if it belongs to one of the $K$ clusters and a bag is labeled as positive bag only if it contains at least one positive instance. The goal of MCIL is to learn one boosting classifier $H_T^k(s_{ij}^k)$ for each of the $K$ clusters. In our clustered data representation settings, each video frame is encoded by a $8\times6$ matrix. If we treat each column vector as an independent instance, all the instances in one bag will form six clusters naturally. A six-cluster MCIL learner can be trained and the overall decision is based on the cluster classifier that gives maximum output:
\begin{equation}
  H_T(s_{ij})=\max_k (H_T^k(s_{ij}))
\end{equation}

Similar to the core of MIL, the posterior probability of bag $i$ is given by,
\begin{equation}
  p_i=\max_j\max_k(p_{ij}^k)
\end{equation}
where $j\in\{1,2,...,N_i\}$ and $k\in\{1,2,...,K\}$. The max function is approximated by a soft-max function,
\begin{equation}
  p_i=g_{jk}(p_{ij}^k)=g_j(g_k(p_{ij}^k))
\end{equation}
Note that the order of soft-max functions is interchangeable, i.e.
\begin{equation}
  g_j(g_k(p_{ij}^k))=g_{jk}(p_{ij}^k)=g_k(g_j(p_{ij}^k))
\end{equation}
Taking GM model as an example, the proof of (19) goes as follows,
\begin{equation}
 \begin{split}
  g_k(g_j(p_{ij}^k)) &= (\frac{1}{K}\sum_k(g_j(p_{ij}^k))^u)^\frac{1}{u} \\
                     &= (\frac{1}{K}\sum_k((\frac{1}{N_i}\sum_j(p_{ij}^k)^u)^\frac{1}{u})^u)^\frac{1}{u} \\
                     &= (\frac{1}{KN_i}\sum_{k,j}(p_{ij}^k)^u)^\frac{1}{u} = g_j(g_k(p_{ij}^k))
 \end{split}
\end{equation}

Finding the strong classifier of a cluster follows a standard boosting procedure and all the classifiers are trained on the same set of BOW instances. However, the weight of instances are updated respectively as per cluster,
\begin{equation}
  \omega_{ij}^k=-\frac{\partial\mathcal{L}}{\partial H_T^k(s_{ij})}
  =-\frac{\partial\mathcal{L}}{\partial p_{i}}\frac{\partial p_{i}}{\partial p_{ij}^k}\frac{\partial p_{ij}^k}{\partial H_T^k(s_{ij})}
\end{equation}
The partial derivative of $\frac{\partial p_{i}}{\partial p_{ij}^k}$ for the GM model is given by,
\begin{equation}
  \frac{\partial p_{i}}{\partial p_{ij}^k}=p_i \frac{(p_{ij}^k)^{(u-1)}}{\sum_{s=1}^{N_i}\sum_{t=1}^{K} (p_{is}^t)^u}
\end{equation}
For the remaining two items in the partial derivative of the weight update expression, $\frac{\partial\mathcal{L}}{\partial p_{i}}$ is the same as in MILboost and $\frac{\partial p_{ij}^k}{\partial H_T^k(s_{ij})}=p_{ij}^k(1-p_{ij}^k)$, which is the derivative $w.r.t$ a sigmoid function.

\section{Experimental Results}
The pain detection system is first tested on the UNBC-McMaster dataset, where video sequences with OPI ratings$\geq3$ are treated as positive samples and those with OPI$=0$ are treated as negative samples. This yielded the same set of 147 sequences from 23 subjects as in \cite{sikka2014classification}. Video sequences are processed by Emotient and the output of single AU Evidence score dataflows are encoded with the compact and clustered AU combination structures separately. These two type of BOW features are used to train MIL (refered to as Compact-MIL) and MCIL (referred to as Clustered-MCIL) learners respectively. Instances in a bag are generated by two temporal aggregation methods, Sc-wind and Ncuts. We set the multiple scaling window size at $30,40,50$ for Sc-wind. The size of segment in a cluster is limited between $21$ and $81$ for Ncuts, and $\sigma_f=0.1,\:\sigma_t=30$ in computing the frame correlation matrix. The GM soft-max function is adopted for the approximation of $\max$ in MIL training. The performance of MIL and MCIL learners are evaluated by accuracy and area under curve (AUC). We use the work in \cite{sikka2014classification}, referred to as MS-MIL, for comparison and the results based on a $10$-fold cross-validation are summarized in Table II.
\begin{table}[ht]
\caption{Comparison of the decoupled framework with MS-MIL \cite{sikka2014classification}} % title of Table
\centering % used for centering table
\begin{tabular}{|c|c|c|c|c|} % centered columns (2 columns)
\hline %inserts double horizontal lines
\multirow{2}{*}{Framework} &
\multicolumn{2}{c|}{Sc-wind} &
\multicolumn{2}{c|}{Ncut}\\ [0.5ex] % inserts table heading
\cline{2-3}\cline{4-5}
 & Accuracy($\%$) & AUC  & Accuracy($\%$)  & AUC \\
\hline % inserts single horizontal line
MS-MIL & 83.7 & & 82.99 & \\
Compact-MIL & 83.96 &0.875 &85.04 &0.9 \\
Clustered-MCIL &85.18 &0.92 &\textbf{86.84} & \textbf{0.94}\\[1.5ex] % [1.5ex] adds vertical space
\hline %inserts single line
\end{tabular}
\label{table:nonlin} % is used to refer this table in the text
\end{table}

The best performance is achieved by the Clustered-MCIL framework in conjunction with Ncuts. The improvement on classifier accuracy may be attributed to the advantage of the decoupled structure. AU coding is a 'hard' problem involving learning a mapping from a very high-dimensional pixel space to the low-dimensional AU space in a complex environment. However, due to the fact that AUs are evoked by certain facial muscular movement, reliable AU coding can be achieved by a robust AFER system if trained on sufficient data. On the other hand, pain is a subjective measure and can be viewed as a latent variable. It will be easier to synthesis similarity from low-dimensional features that are highly correlated to pain under the impact of problem uncertainty. In addition, the improvement on AUC between Clustered-MCIL and Compact-MIL could be attributed to the feature sparsity from the clustered representation, which not only increases the margin on features but also follows more naturally to a human coder's decisions.

Next, we employ the Clustered-MCIL settings with GM approximation to train a pain detector on UNBC-McMaster dataset and test it on video sequences from selected patients in Wilkie's dataset. Each test sequence is divided into $30$ subsequences and each subsequence has a duration of $20$ seconds. Three human expert coders performed FACS coding on each subsequence and pain is identified if any pain-related AU combination is detected by at least two human expert coders in the original research. However, since the human coding does not reveal pain intensity information and coders do not always agree with each other, this ground truth is more suitable for qualitative analysis. $393$ subsequences from $27$ patients are selected for this experiment, where at least $50\%$ of each subsequence is analyzable by Emotient. A subsequence is considered as positive (pain) if AU combinations are coded by at least two coders (more credible) or only one coder (less credible). A subsequence is negative (no pain) if no AU combination is scored by any coder.  As a result, $82$ subsequences are identified as positive by at least two coders, $121$ subsequences are identified as negative by only one coder, and the rest $190$ subsequences are identified as negative samples. The automated system is used as an independent coder and it checks the consistency between machine prediction and human coder decisions and the results are summarized in Table III. Note that the system has no prior knowledge about the test dataset. In general, observe that the decisions of automated system are highly correlated with that of a majority of human coders on both pain and no pain videos. Additionally, a $68.6\%$ consistency rate is observed between the system and the only coder, however this is more likely due to the ambiguity in less credible videos rather than to the accuracy degradation. As a result, the system shows its potential in pain assessment for long videos under clinical settings and we will conduct further tests with focus on patient monitoring in future research.
\begin{table}
\caption{Comparison of machine prediction with human coder decision on subsequences in Wilkie's dataset\cite{wilkie1995facial}} % title of Table
\newcommand{\tabincell}[2]{\begin{tabular}{@{}#1@{}}#2\end{tabular}}
\centering
\begin{tabular}{|c|c|c|c|}\hline
\tabincell{c}{Subsequences\\Label}&\tabincell{c}{Pain\\($2+$ coders\\ scored)}&\tabincell{c}{Pain\\($1$ coder\\ scored)}&\tabincell{c}{No Pain\\($3$ coders\\ agreed)}\\[0.5ex]\hline
Human Coder & $82$& $121$&$190$\\[0.5ex]\hline
Automated System & $68$ & $83$ &$169$\\[0.5ex]\hline
Consistency Rate &$82.9\%$ &$68.6\%$ &$88.9\%$\\[0.5ex]\hline
\end{tabular}
\end{table}

\section{Conclusion}
This paper proposed and investigated the performance of automated pain detection via spontaneous facial expressions in the context of clinical applications. The proposed framework mimics the decision strategy of FACS-certified human coder by following the procedures of \emph{Facial Expression} $\rightarrow$\emph{AU}$\rightarrow$\emph{AU Combination}$\rightarrow$\emph{Pain}. To address the challenge in accessing sufficient pain annotated video data, we proposed a decoupled structure for automated pain detection task: 1) the AU coding from facial expressions takes full advantage of the AFER development, 2) pain detection is based on simply low-dimensional features and handled by MIL as a weakly supervised problem. The proposed system not only demonstrates improvement on existing state-of-the-art work, but also shows adaptivity on trans-dataset learning and long video analysis. In future work, we will conduct comprehensive test on new pain oriented videos datasets and investigate practical methodologies that facilitate clinical pain analysis using our system.


\section*{Acknowledgment}

This research and publication was made possible in past by Grant Number P30 NR010680-S1 from the National Institutes of Health, National Institute of Nursing Research. Its contents are solely the responsibility of the authors and do not necessarily represent the official views of the National Institute of Nursing Research.



%\bibliographystyle{IEEEtran}

\bibliographystyle{plain}
\bibliography{MIL_reference}

%\begin{thebibliography}{1}
%\end{thebibliography}
%
%

\end{document}